\documentclass[conference]{IEEEtran}
\IEEEoverridecommandlockouts
\usepackage{cite}
\usepackage{amsmath,amssymb,amsfonts}
\usepackage{algorithmic}
\usepackage{graphicx}
\usepackage{todonotes}
\usepackage{textcomp}
\usepackage{xcolor}
\usepackage{amsmath}
\usepackage{url}
\usepackage{breakurl}
\usepackage[breaklinks]{hyperref}
\newcommand{\SO}{\ensuremath{\mathsf{SO(3)}}}
\newcommand{\SE}{\ensuremath{\mathsf{SE(3)}}}
\renewcommand{\Re}{\ensuremath{\mathbb{R}}}
\newcommand{\so}{\ensuremath{\mathfrak{so}(3)}}

\DeclareMathOperator{\atantwo}{atan2}

\DeclareMathAlphabet{\pazocal}{OMS}{zplm}{m}{n}

\newcommand{\Bs}{\pazocal{B}}

\newcommand{\Es}{\pazocal{E}}

\newcommand{\Vs}{\pazocal{V}}

\def\BibTeX{{\rm B\kern-.05em{\sc i\kern-.025em b}\kern-.08em
    T\kern-.1667em\lower.7ex\hbox{E}\kern-.125emX}}

\begin{document}

\title{Aerial Gym – Isaac Gym Simulator for Aerial Robots \\
\thanks{This material was supported by the AFOSR Award No. FA8655-21-1-7033. All authors are affiliated with the Norwegian University of Science and Technology (NTNU), Trondheim, Norway.}
}

\author{\IEEEauthorblockN{Mihir Kulkarni}
% Trondheim, Norway \\
% mihir.kulkarni@ntnu.no}
\and
\IEEEauthorblockN{Theodor J. L. Forgaard}
% Trondheim, Norway \\
% tjforgaa@stud.ntnu.no}
\and
\IEEEauthorblockN{Kostas Alexis}
% Trondheim, Norway \\
% konstantinos.alexis@ntnu.no}
}

\maketitle

\begin{abstract}
Developing learning-based methods for navigation of aerial robots is an intensive data-driven process that requires highly parallelized simulation. The full utilization of such simulators is hindered by the lack of parallelized high-level control methods that imitate the real-world robot interface. Responding to this need, we develop the Aerial Gym simulator that can simulate millions of multirotor vehicles parallelly with nonlinear geometric controllers for the Special Euclidean Group SE(3) for attitude, velocity and position tracking. We also develop functionalities for managing a large number of obstacles in the environment, enabling rapid randomization for learning of navigation tasks. In addition, we also provide sample environments having robots with simulated cameras capable of capturing RGB, depth, segmentation and optical flow data in obstacle-rich environments. This simulator is a step towards developing a -- currently missing -- highly parallelized aerial robot simulation with geometric controllers at a large scale, while also providing a customizable obstacle randomization functionality for navigation tasks. We provide training scripts with compatible reinforcement learning frameworks to navigate the robot to a goal setpoint based on attitude and velocity command interfaces. Finally, we open source the simulator and aim to develop it further to speed up rendering using alternate kernel-based frameworks in order to parallelize ray-casting for depth images thus supporting a larger number of robots.
\end{abstract}


\section{Introduction}
In recent years, aerial robots have gained significant attention in various applications, ranging from search and rescue missions to automated site inspections.
These environments can present complex geometries and may present challenges to the perception systems owing to darkness, airborne particles, geometric self-similarities, etc. Due to the highly complex nature of the problem, there is a renewed interest in utilizing learning-based navigation methods to operate in these settings.
% A crucial factor ensuring the efficient operation of these aerial robots is the development of robust learning-based navigation methods. 
The development of such methods is an intensive data-driven process, necessitating highly parallelized simulation environments. Despite the existence of powerful simulators, the full potential of these tools remains untapped due to the absence of parallelized control methods that imitate the real-world robot interface and allow users to provide high-level commands to the robot.

\begin{figure}[t!]
\includegraphics[width=\columnwidth]{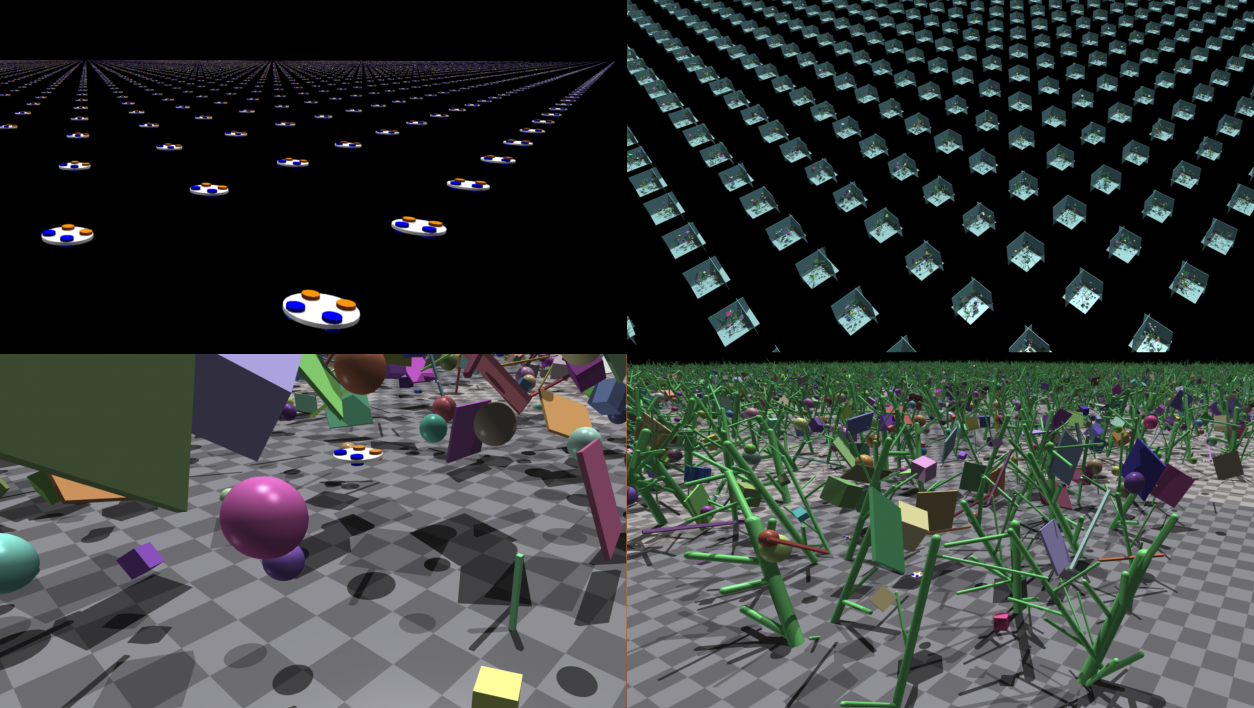}
\vspace{-2ex}
\caption{Visualization of the Aerial Gym simulator with multiple simulated multirotor robots. Instances show -- in clockwise order -- the simulation of the robots in obstacle-free environments, a zoomed-out view of separated box-like environments, as well as cluttered environments for navigation consisting of randomly distributed obstacles of different shapes.}
\label{fig:intro_image}
\vspace{-2ex}
\end{figure}

We address this gap by harnessing the capabilities of NVIDIA Isaac Gym to design an aerial robot simulator that enables the parallel simulation of thousands of multirotor vehicles. We integrate nonlinear geometric controllers for the Special Euclidean Group \SE~\cite{lee2011control} to achieve accurate attitude and velocity tracking in the vehicle frame, or position tracking in the inertial frame, while taking advantage of GPU parallelization. We rely on the underlying physics interface to simulate forces, torques, and collisions, while modeling the robot as a rigid body.

% Moreover, we develop interfaces for efficiently managing a large set of objects in the environment. These obstacles, defined using URDFs, facilitate rapid randomization for learning in cluttered environments. 
Moreover, we develop interfaces for efficiently managing a large set of objects in the environment, defined using Universal Robot Description Format (URDF) files. This facilitates rapid randomization of the environment for learning in cluttered settings. Our simulator provides sample environments containing robots equipped with simulated cameras capable of capturing RGB, depth, segmentation, and optical flow data at thousands of frames per second (a total of $\sim1800~\textrm{fps}$ for $2^{10}$ robots). These powerful tools and simulators mark a significant step towards the development of a highly parallelized aerial robot simulation environment with geometric controllers at a large scale, coupled with a customizable obstacle randomization functionality. We release our simulator as open-source software at \url{https://github.com/ntnu-arl/aerial_gym_simulator} and shall continue its development by leveraging alternative kernel-based frameworks to parallelize ray-casting for depth images, ultimately supporting simulation of an even larger number of robots. We will also release example scripts that show the robots interfaced with commonly used learning frameworks (e.g., cleanRL~\cite{huang2022cleanrl} or rl-games~\cite{rl-games2021}) to learn to reach a target setpoint using attitude control in an obstacle-free environment.

In the remainder of this paper, Section~\ref{sec:RelatedWork} presents the related work, followed by the design of the proposed simulator for aerial robots in Section~\ref{sec:SimulatorDesign}. Sections~\ref{sec:controllers} and~\ref{sec:asset_manager} detail the overall design, the parallelized geometric controller and the obstacle asset management functionality respectively. Section~\ref{sec:Benchmarks and Evaluation} benchmarks the performance of the simulator, followed by conclusions in Section~\ref{sec:Conclusions}.

\section{Related Work}\label{sec:RelatedWork}
Several simulators have been developed to simulate aerial robots for a variety of tasks such as navigation, mapping, and control. RotorS~\cite{Furrer2016rotors} is a Gazebo-based~\cite{koenig2004gazebosim} simulator that provides a variety of multirotors with RGB-D sensors. Airsim~\cite{airsim2017fsr} is a photo-realistic simulator built on Unreal Engine. The simulator supports both hardware- and software-in-the-loop simulations for a limited number of simulated robots. Flightmare~\cite{song2020flightmare} offers the functionality to simulate a large number of robots in parallel. However, the robot dynamics simulated by Flightmare are calculated on parallel threads on the CPU, limiting the number of robots that can be simulated. Flightmare uses the Unity rendering engine, allowing high-fidelity graphics simulation.

NVIDIA's Isaac Gym~\cite{makoviychuk2021isaac} provides GPU-accelerated highly parallelized simulation functionality for robot learning tasks. This simulator is being extensively used to simulate articulated and multi-linked robots~\cite{rudin2022learning}. Some simplified simulation environments for aerial robots have also been developed to work with it~\cite{makoviychuk2021isaac}, however, the models either lack fidelity, or they only provide interfaces to command motor forces but ignore the effect of torque generated by the motor on the body. They also do not support any other higher-level reference tracking interfaces. Accordingly, to provide the simulation capability exploiting the GPU, our simulator is built upon the Isaac Gym simulator. Then, we further provide GPU-based geometric attitude and velocity controllers thus supporting a wider range of control inputs enabling the simulator's utility to a larger set of use cases and the capability to train for real-world deployments with potentially reduced sim-to-real gap.

\section{Simulator Design}\label{sec:SimulatorDesign}

\begin{figure}[t!]
\includegraphics[width=\columnwidth]{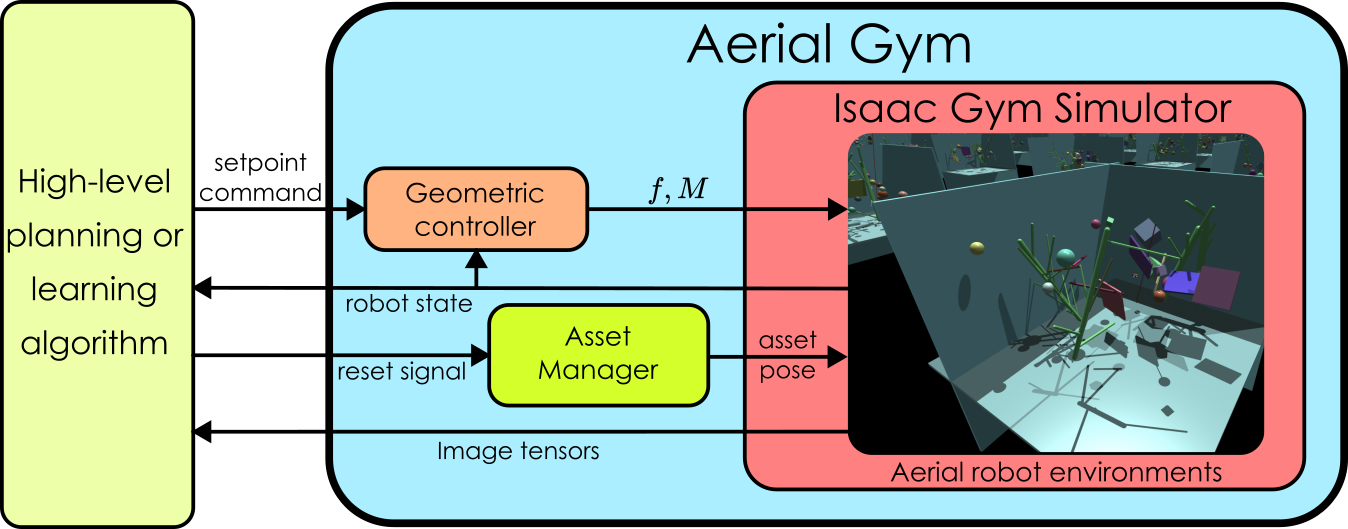}
\caption{Block diagram of the Aerial Gym simulator with the components to control the simulated robots and manipulate and randomize the simulated obstacles (also called assets) in multiple parallel environments.}
\label{fig:simulator_design}
\vspace{-4ex}
\end{figure}

We build the proposed Aerial Gym simulator utilizing the tensor-based parallelization provided by NVIDIA Isaac Gym simulator~\cite{makoviychuk2021isaac}. We design our simulator with the appropriate interfaces to imitate standardized reinforcement learning environments~\cite{brockman2016openaigym} in order to facilitate easy extension of commonly-used learning-based algorithms for robot navigation. We design a generalized asset management class, that allows a user to load URDF files describing obstacles from a folder structure. We define an asset - in line with NVIDIA Isaac Gym - to represent a mesh entity in simulation that may contain links, joints, and other physical properties. In our case, assets are considered to be represented by URDF files, however, extending this to other supported formats is trivial. The selection and loading of these files happen in a randomized fashion per environment, where a predefined number of different obstacle meshes per custom-defined class of obstacles are picked randomly to be included in a simulation environment. This allows various environments to have sufficient diversity and prevent the reuse of identical obstacle meshes across environments. The robots simulated in obstacle-rich environments can also be equipped with camera sensors capable of capturing RGB, depth, segmentation, and optical flow images. The position and orientation of the camera sensors with respect to the robot are configurable and may be randomized. We aggregate images into a consolidated tensor and provide direct access to this tensor. In addition, we adapt nonlinear geometric controllers for aerial robots and -- importantly -- parallelize the controllers to be run on the GPU. This allows for the capability of providing high-level input commands to the robots, relying on interfaces that are available on commonly used flight controllers for multirotor aerial vehicles~\cite{px4}. The structure of the simulator is shown as a block diagram in Figure~\ref{fig:simulator_design}, where the interaction between various modules is highlighted. A high-level planner or a learning-based framework can access the robot state which includes the position, orientation, linear velocity, and angular velocity of the robot. Access is also provided to the image tensors from the simulated sensors onboard the robot. Data from these sensors can be utilized by user-provided high-level planning or learning methods for navigation tasks or simulated data collection. In addition, access to the state of each obstacle is made available to the user as privileged information for training learning-based methods. As an additional contribution, we provide example scripts to procedurally generate URDF models of simplified multi-linked tree-like objects that can directly be added to the simulator for training learning-based methods to navigate forest-like environments. These procedurally generated trees have configurable length, diameters and branching factors, allowing the users to randomize the generated meshes to learn collision avoidance in diverse sets of environments.

\section{Parallelized Geometric Controller on \SE}\label{sec:controllers}
We develop the parallelized controller on \SE\, based on the work of~\cite{lee2011control}. The inertial reference frame is denoted as $\Es$ with basis vectors \{$\vec e_1$, $\vec e_2$, $\vec e_3$\} and a body-fixed frame $\Bs$ with basis vectors \{$\vec b_1$, $\vec b_2$, $\vec b_3$\}. We additionally define a vehicle frame $\Vs$ with the basis vectors \{$\vec v_1$, $\vec v_2$, $\vec v_3$\} that is yaw-aligned with the body-fixed frame, and having the $x-y$ plane parallel to the inertial frame. We define
\begin{tabular}{lp{5.6cm}}
{$m\in\Re$} & the total mass\\
{$g\in\Re^{3}$} & the gravity vector \\
{$\psi\in\Re$} & the current yaw angle of the robot\\
{$\phi_d\in\Re$} & the reference roll angle\\
{$\theta_d\in\Re$} & the reference pitch angle\\
{$\Dot{\psi}_d\in\Re$} & the reference yaw rate\\
{$J\in\Re^{3\times 3}$} & the inertia matrix with respect to the body-fixed frame\\
{$R\in\SO$} & the rotation matrix from the body-fixed frame to the inertial frame\\
{$R_d\in\SO$} & the rotation matrix from desired body-fixed frame to the inertial frame\\
{$\Omega\in\Re^3$} & the angular velocity in the body-fixed frame\\
{$\Omega_d\in\Re^3$} & the desired angular velocity in the desired body-fixed frame\\
{$v\in\Re^3$} & the velocity vector of the center of mass in the vehicle frame\\
{$v_d\in\Re^3$} & the desired velocity vector of the center of mass in the vehicle frame\\
{$f\in\Re$} & the total thrust magnitude along the $-\vec b_3$ axis\\
{$M\in\Re^3$} & the total moment vector in the body-fixed frame.\\
\end{tabular}

The position of the aerial robot is defined by the location of the center of mass and the attitude is expressed with respect to the inertial frame. Due to the underactuated nature of quadrotors (and most other multirotor systems), asymptotic output tracking of both attitude and position is not possible. Two flight modes are detailed in this section. We consider the control of the robot using either a) the attitude-controlled flight mode or b) velocity-controlled flight mode, and provide adapted controllers for the same.

\subsection{Attitude-Controlled Flight Mode}\label{sec:attitude_control}
To imitate widely available real-world control interfaces, we consider $\phi_d$, $\theta_d$, $\Dot{\psi}_d$ and $f$ as command inputs to the attitude controller and calculate the desired body-fixed frame orientation, $R_d$, and the desired angular velocity, $\Omega_d$, as below:

\begin{align}
    R_d &= R_z(\psi) R_y(\theta_d) R_x(\phi_d)\\
    \Omega_d &= \begin{pmatrix} 1 & 0 & -\sin{\theta_d} \\ 0 & \cos{\phi_d} & \sin{\phi_d} \cos{\theta_d} \\ 0 & -\sin{\phi_d} & \cos{\phi_d}\cos{\theta_d} \end{pmatrix} \begin{pmatrix} 0 \\ 0 \\ \Dot{\psi}_d \end{pmatrix}
\end{align}
where $\mathbf{R}_j(\eta)~(j=x,y,z)$ denotes the rotation matrix for a rotation around the $j$-axis by $\eta$ degrees. The attitude tracking error $e_R \in \Re$ is expressed as:

\begin{equation}
    e_R = \frac{1}{2} (R_d^TR - R^T R_d)^\vee,
\end{equation}
where the \textit{vee map} $^\vee:\so\rightarrow\Re^3$ maps skew-symmetric matrices in $\Re^3$. The angular velocity error $e_\Omega \in \Re^3$ is expressed as

\begin{equation}
    e_\Omega = \Omega - R^T R_d \Omega_d.
\end{equation}
The nonlinear controller for the attitude-controlled flight mode is defined as
 
\begin{equation}
    M = -k_R e_R - k_\Omega e_\Omega + \Omega \times J \Omega,
\end{equation}
where $k_R$ and $k_\Omega$ are diagonal matrices with positive entries. The final term from the corresponding equation in~\cite{lee2011control} is dropped, similar to the implementation in~\cite{Furrer2016rotors}.

\subsection{Velocity-Controlled Flight Mode}\label{sec:velocity_control}
Similarly, a nonlinear controller for the velocity-controlled flight mode is utilized. An arbitrary velocity tracking command ${}^{\Vs}v_d \in \Re^3 $ is given, with the desired yaw-rate $\Dot{\psi}_d$. The commanded acceleration vector can be calculated from the velocity tracking error as:

\begin{equation}
    a_d = k_v (v_d - v),
\end{equation}
where $k_v$ is a diagonal matrix with positive entries. From this, the total thrust and the commanded tilt angles can be deduced as:
\begin{align}
    f &= (a_d + m g) \cdot {}^{\Vs}\vec{b_3} \\
    \phi_d &= \atantwo(-a_{d,y}, \sqrt{a^2_{d,x} + a^2_{d,z}}) \\
    \theta_d &= \atantwo(a_{d,x}, a_{d,z}),
\end{align}
where $a_{d,j}~(j=x,y,z)$ is the $j-$component of $a_d$, and ${}^{\Vs}\vec{b_3}$ is the transformed coordinates of $\vec{b_3}$ in $\Vs$. The reference tilt angles and reference yaw rate can then be tracked by the low-level attitude controller described above. The force $f$ and torque $M$ are applied to the simulated rigid-body robot in Isaac Gym using the underlying physics engine.

% We assume an arbitrary attitude ($R_d(t) \in \SO$) and angular velocity tracking commands ($\Omega_d(t) \in \Re^3$) are provided as a function of time. We consider that the commands are expressed in the vehicle frame. The attitude tracking error $e_R \in \Re$ is expressed as

% \begin{equation}
%     e_R = \frac{1}{2} (R_d^TR - R^T R_d)^\vee,
% \end{equation}
% where the \textit{vee map} $^\vee:\so\rightarrow\Re^3$ maps skew-symmetric matrices in $\Re^3$. The angular velocity error $e_\Omega \in \Re^3$ is expressed as

% \begin{equation}
%     e_\Omega = \Omega - R^T R_d \Omega_d.
% \end{equation}
% The nonlinear controller for the attitude controlled flight mode is defined as

% \begin{equation}
%     M = -k_R e_R - k_\Omega e_\Omega + \Omega \times J \Omega.
% \end{equation}
% The last term can be dropped in practice as it does not appear to make a significant difference.

\section{Asset Manager}\label{sec:asset_manager}

\begin{figure}
    \centering
    \includegraphics[width=\columnwidth]{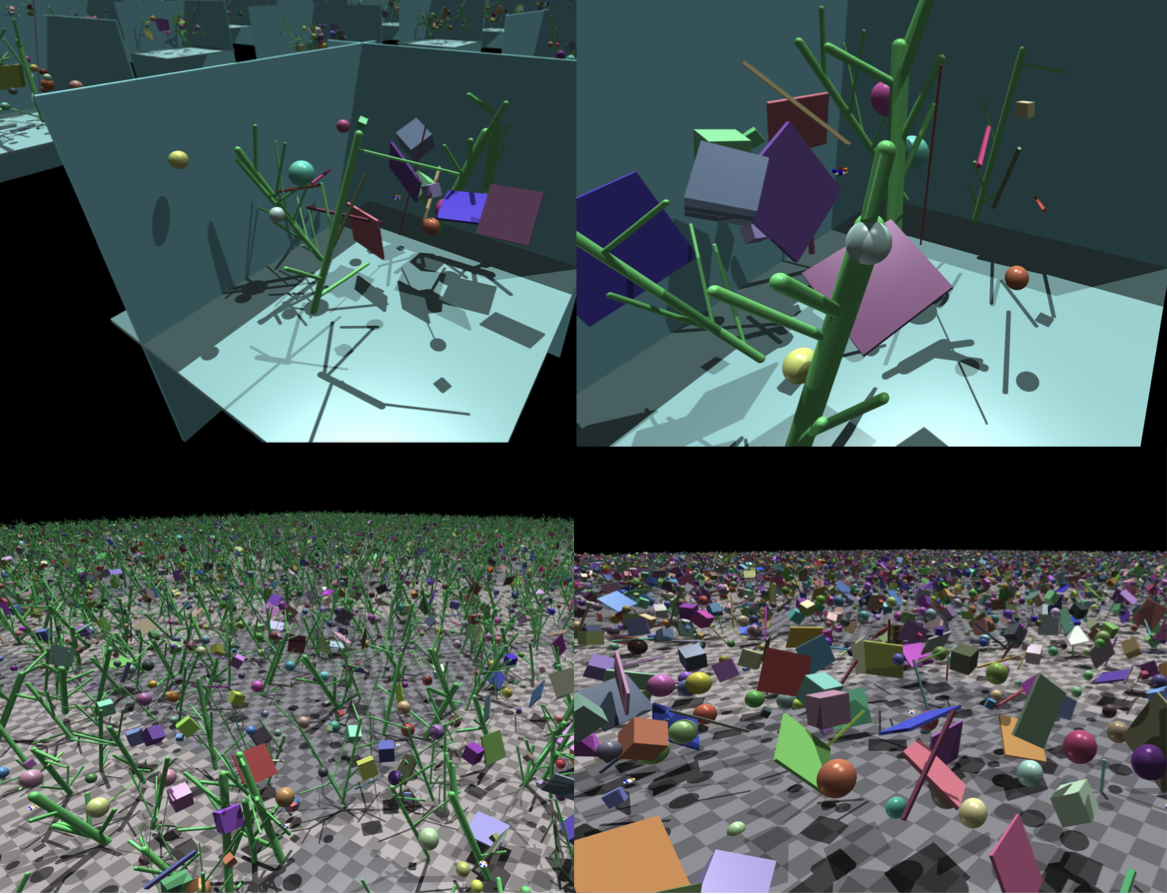}
    \caption{Examples of different environments configured. The top row shows bounded environments, with the front, top, and left walls removed for visualization. The bottom row shows different configurations for multiple environments without walls.}
    \label{fig:environment_examples}
\end{figure}

To robustly train navigation policies using exteroceptive sensor data, environments containing obstacles and general clutter need to be created. Since the Isaac Gym simulator~\cite{makoviychuk2021isaac} provides tensor-level access to the position, orientation, and linear/angular velocities of each object in the simulator, we exploit this feature to build classes to easily modify each environment to randomize the obstacles. The simulator allows us to separately consider each individual environment, where collisions can be separated and masked between various entities. However, modifying each obstacle in the environment can be a cumbersome process. To avoid this, we develop a tensor-based asset manager to easily configure, randomize and manipulate the objects that are created in the simulator. Various objects are categorized into user-defined classes (e.g., thin obstacles, large trees, geometric shapes, etc.) based on their use case. Multiple URDF files for each class are stored in a folder structure named after the name of the associated class of objects. A configuration file describes the number of obstacles from each class to be created in each environment. The asset manager randomly picks (with replacement) the required number of URDF files separately for each environment. This allows the user to populate a directory with a large number of models, of obstacles with various types, shapes, and sizes, ensuring that there is randomization across various environments. This prevents learning methods from overfitting to specific types of obstacles. The asset manager is also utilized when the environments are reset as it samples a random position and orientation for each obstacle in the environment that is to be reset. The minimum and maximum bounds for the position can be set as a fraction of the environment bounds, while the orientation bounds can be set as numerical values corresponding to euler angles. The randomization in specific dimensions can be prevented by assigning a  constant value to specific dimensions for chosen asset types. This allows placing obstacles at predefined locations across environments. While we provide only randomization for position and orientation for static obstacles, extending it to dynamic obstacles is trivial. The asset manager uses the NVIDIA Isaac Gym API and provides an easier interface to easily add segmentation labels to different classes of obstacles. Some examples of randomized environments generated with the help of the asset manager are shown in Figure~\ref{fig:environment_examples}.

% In addition, the asset manager is able to randomly pick out a URDF file from a set of files defining the collision and visual structure of an object belonging to a class to ensure that the same object is not used across environments. We develop configurable classes to control the number of instances of the specific classes of objects, which of the assets states should be kept fixed or randomized, define the minimum and maximum values of the position, euler angles, and configure the segmentation mask of each link or object for generating data using the segmentation camera. We currently only support static objects and our manager include randomization of position and orientation. However, this method can be further extended to support moving objects with minimal effort.

% \begin{itemize}
%     \item env manager functionality, locked states, min,max states
%     \item randomization, random number of obstacles
% \end{itemize}

\section{Benchmarks and Evaluation}\label{sec:Benchmarks and Evaluation}
To benchmark the performance of the simulator we simulate environments (without obstacles) with $2^{17}$ robots and command a set attitude or velocity. The parallelized simulation, including that of the implemented controllers, allows the Aerial Gym simulator to simulate an aggregate of over $3.8\times10^6$ steps per second, with each step corresponding to $10~\textrm{ms}$ of simulated time. Effectively, the data generated for the robot simulation from the simulator is equivalent to obtaining a speedup of $3.8\times10^4$ as compared to a real-time run on a single robot. However, the addition of camera sensors restricts the number of robots that can be used. The Isaac Gym API also does not allow direct access to a consolidated camera tensor leading to an overhead to read each image separately. We simulate $2^{10}$ robots with a depth camera with a resolution of $270\times480$ pixels and can render a total of up to $1,800$ frames per second. Each of these tests was performed on a workstation with $2\times$ NVIDIA RTX $3090$ GPUs and an AMD Ryzen Threadripper PRO $3975$WX CPU with $32$-Cores.

\section{Conclusion}\label{sec:Conclusions}
In this work, we presented the Aerial Gym simulator for aerial robots, capable of parallelly simulating thousands of flying robots. We adapted a nonlinear geometric controller to work with parallelly simulated robots to provide high-level interfaces that allow the imitation of real-world control interfaces. Asset management functionality is packaged to allow randomization of obstacles across environments and the creation of custom user-defined obstacle classes for effective handling of each kind of obstacle. Finally, we open-source our simulator to enable the aerial robotics community to leverage the benefit of parallelized simulation frameworks. Future developments on this simulator include the utilization of alternate kernel-based ray-casting frameworks allowing a rapid speedup in rendering depth images with a higher number of robots. The software for the simulator is made available at \url{https://github.com/ntnu-arl/aerial_gym_simulator}.

%\cite{IEEEexample:miscgermanreg}
\bibliographystyle{IEEEtran}
\bibliography{bib}

\begin{thebibliography}{10}
\providecommand{\url}[1]{#1}
\csname url@rmstyle\endcsname
\providecommand{\newblock}{\relax}
\providecommand{\bibinfo}[2]{#2}
\providecommand\BIBentrySTDinterwordspacing{\spaceskip=0pt\relax}
\providecommand\BIBentryALTinterwordstretchfactor{4}
\providecommand\BIBentryALTinterwordspacing{\spaceskip=\fontdimen2\font plus
\BIBentryALTinterwordstretchfactor\fontdimen3\font minus
  \fontdimen4\font\relax}
\providecommand\BIBforeignlanguage[2]{{%
\expandafter\ifx\csname l@#1\endcsname\relax
\typeout{** WARNING: IEEEtran.bst: No hyphenation pattern has been}%
\typeout{** loaded for the language `#1'. Using the pattern for}%
\typeout{** the default language instead.}%
\else
\language=\csname l@#1\endcsname
\fi
#2}}

\bibitem{lee2011control}
T.~Lee, M.~Leok, and N.~H. McClamroch, ``Control of complex maneuvers for a
  quadrotor uav using geometric methods on se(3),'' 2011.

\bibitem{huang2022cleanrl}
\BIBentryALTinterwordspacing
S.~Huang, R.~F.~J. Dossa, C.~Ye, J.~Braga, D.~Chakraborty, K.~Mehta, and J.~G.
  Araújo, ``Cleanrl: High-quality single-file implementations of deep
  reinforcement learning algorithms,'' \emph{Journal of Machine Learning
  Research}, vol.~23, no. 274, pp. 1--18, 2022. [Online]. Available:
  \url{http://jmlr.org/papers/v23/21-1342.html}
\BIBentrySTDinterwordspacing

\bibitem{rl-games2021}
D.~Makoviichuk and V.~Makoviychuk, ``rl-games: A high-performance framework for
  reinforcement learning,'' \url{https://github.com/Denys88/rl\_games}, May
  2021.

\bibitem{Furrer2016rotors}
\BIBentryALTinterwordspacing
F.~Furrer, M.~Burri, M.~Achtelik, and R.~Siegwart, \emph{Robot Operating System
  (ROS): The Complete Reference (Volume 1)}.\hskip 1em plus 0.5em minus
  0.4em\relax Cham: Springer International Publishing, 2016, ch. RotorS---A
  Modular Gazebo MAV Simulator Framework, pp. 595--625. [Online]. Available:
  \url{http://dx.doi.org/10.1007/978-3-319-26054-923}
\BIBentrySTDinterwordspacing

\bibitem{koenig2004gazebosim}
N.~Koenig and A.~Howard, ``Design and use paradigms for gazebo, an open-source
  multi-robot simulator,'' in \emph{2004 IEEE/RSJ International Conference on
  Intelligent Robots and Systems (IROS) (IEEE Cat. No.04CH37566)}, vol.~3,
  2004, pp. 2149--2154 vol.3.

\bibitem{airsim2017fsr}
\BIBentryALTinterwordspacing
S.~Shah, D.~Dey, C.~Lovett, and A.~Kapoor, ``Airsim: High-fidelity visual and
  physical simulation for autonomous vehicles,'' in \emph{Field and Service
  Robotics}, 2017. [Online]. Available: \url{https://arxiv.org/abs/1705.05065}
\BIBentrySTDinterwordspacing

\bibitem{song2020flightmare}
Y.~Song, S.~Naji, E.~Kaufmann, A.~Loquercio, and D.~Scaramuzza, ``Flightmare: A
  flexible quadrotor simulator,'' in \emph{Conference on Robot Learning}, 2020.

\bibitem{makoviychuk2021isaac}
V.~Makoviychuk, L.~Wawrzyniak, Y.~Guo, M.~Lu, K.~Storey, M.~Macklin,
  D.~Hoeller, N.~Rudin, A.~Allshire, A.~Handa, and G.~State, ``Isaac gym: High
  perforemance gpu-based physics simulation for robot learning,'' 2021.

\bibitem{rudin2022learning}
N.~Rudin, D.~Hoeller, P.~Reist, and M.~Hutter, ``Learning to walk in minutes
  using massively parallel deep reinforcement learning,'' 2022.

\bibitem{brockman2016openaigym}
G.~Brockman, V.~Cheung, L.~Pettersson, J.~Schneider, J.~Schulman, J.~Tang, and
  W.~Zaremba, ``Openai gym,'' \emph{arXiv preprint arXiv:1606.01540}, 2016.

\bibitem{px4}
L.~Meier, D.~Honegger, and M.~Pollefeys, ``Px4: A node-based multithreaded open
  source robotics framework for deeply embedded platforms,'' in \emph{2015 IEEE
  International Conference on Robotics and Automation (ICRA)}, 2015, pp.
  6235--6240.

\end{thebibliography}

%\begin{thebibliography}{00}
%\bibitem{b1} G. Eason, B. Noble, and I. N. Sneddon, ``On certain integrals of Lipschitz-Hankel type involving products of Bessel functions,'' Phil. Trans. Roy. Soc. London, vol. A247, pp. 529--551, April 1955.
%\bibitem{b2} J. Clerk Maxwell, A Treatise on Electricity and Magnetism, 3rd ed., vol. 2. Oxford: Clarendon, 1892, pp.68--73.
%\bibitem{b3} I. S. Jacobs and C. P. Bean, ``Fine particles, thin films and exchange anisotropy,'' in Magnetism, vol. III, G. T. Rado and H. Suhl, Eds. New York: Academic, 1963, pp. 271--350.
%\bibitem{b4} K. Elissa, ``Title of paper if known,'' unpublished.
%\bibitem{b5} R. Nicole, ``Title of paper with only first word capitalized,'' J. Name Stand. Abbrev., in press.
%\bibitem{b6} Y. Yorozu, M. Hirano, K. Oka, and Y. Tagawa, ``Electron spectroscopy studies on magneto-optical media and plastic substrate interface,'' IEEE Transl. J. Magn. Japan, vol. 2, pp. 740--741, August 1987 [Digests 9th Annual Conf. Magnetics Japan, p. 301, 1982].
%\bibitem{b7} M. Young, The Technical Writer's Handbook. Mill Valley, CA: University Science, 1989.
%\end{thebibliography}

\end{document}